A new automatic approach to seed image analysis: From acquisition to segmentation

Vale A.M.P.G.[1], Ucchesu M.[2*], Di Ruberto C.[3], Loddo A.[3], Soares J.M.[4], Bacchetta G.[2,5]

[1] Escola Agrícola de Jundiaí (EAJ), Universidade Federal do Rio Grande do Norte (UFRN), Rodovia RN 160, Km 03, CEP 59280-000 Macaíba (RN), Brasil.
[2] Centro Conservazione Biodiversità (CCB), Dipartimento di Scienze della Vita e dell'Ambiente (DiSVA), Università degli Studi di Cagliari, Viale S. Ignazio da Laconi 13, 09123 Cagliari, Italy.
[3] Dipartimento di Matematica e Informatica, Università degli Studi di Cagliari, Via Ospedale 72, 09124 Cagliari, Italy.
[4] Universidade Federal do Rio Grande do Norte (UFRN), Rodovia RN 160, Km 03, CEP 59280-000 Macaíba (RN), Brasil.
[5] Hortus Botanicus Karalitanus (HBK), Università degli Studi di Cagliari, Viale S. Ignazio da Laconi 11, 09123 Cagliari, Italy.

Corresponding author: Mariano Ucchesu - email: marianoucchesu@gmail.com

# Abstract

Recently, image analysis has begun to be widely applied in the field of plant sciences. It offers a new tool for classifying vascular plant species based on the morphological and colorimetric features of the seeds, and has made significant contributions in systematic studies.

However, in order to extract the morphological and colorimetric features, it is necessary to segment the image containing the samples to be analysed. This stage represents one of the most challenging steps in image processing, as it is difficult to separate uniform and homogeneous objects from the background. In this paper, we present a new, open source plugin for the automatic segmentation of an image of a seed sample. This plugin was written in Java to allow it to work with ImageJ open source software.

The new plugin was tested on a total of 3,386 seed samples from 120 species belonging to the Fabaceae family. Digital images were acquired using a flatbed scanner. In order to test the efficacy of this approach in terms of identifying the edges of objects and separating them from the background, each sample was scanned using four different hues of blue for the background, and a total of 480 digital images were elaborated.

The performance of the new plugin was compared with a method based on double image acquisition (with a black and white background) using the same seed samples, in which images were manually segmented using the Core ImageJ plugin. The results showed that the new plugin was able to segment all of the digital images without generating any object detection errors. In addition, the new plugin was able to segment images within an average of 0.02 s, while the average time for execution with the manual method was 63 s. This new open source plugin is proven to be able to work on a single image, and to be highly efficient in terms of time and segmentation when working with large numbers of images and a wide diversity of shapes.

[1]https://github.com/andrealoddo/blueBackgroundSeedsSegmenter

## 1. Introduction

In recent years, technological progress and advances in computer science have enabled the application of computer vision in many fields of life science (Chaki and Dey, 2019; Lind, 2012; Scherer, 2020; Sun, 2016). Computer vision is a technology that can be used to identify objects and extract quantitative information (Chaki and Dey, 2019; Sun, 2016). The reliability of this technique lies in its ability to quickly provide objective and repeatable data; it is also a non-destructive technique, and is applicable to a wide range of domains (Chaki and Dey, 2018). Image analysis is now widely applied in the field of plant sciences; it offers a new tool that is able to classify vascular plant species based on the morphological and colorimetric features of the seeds, and has made significant contributions in systematic studies. In particular, image processing techniques have been successfully used in the completely automatic identification in biology (Campanile et al. 2019) an in plant species, based on leaf recognition, for the purposes of cataloguing and preserving plants (Di Ruberto and Putzu, 2014; Putzu et al., 2016). Image analysis is also widely used in systematic botany for the characterisation of germplasm (Bacchetta et al., 2011a; Bacchetta et al., 2011b; Frigau et al., 2019; Grillo et al., 2013; Lo Bianco et al., 2015; Lo Bianco et al., 2017a; Lo Bianco et al., 2017b; Mattana et al., 2008; Murru et al., 2019; Pinna et al., 2014; Sarigu et al., 2019), in archaeobotany for the identification of unknown plant seed materials (Bouby et al., 2013; Orrù et al., 2013; Sabato et al., 2015; Terral et al., 2009; Ucchesu et al., 2016; Ucchesu et al., 2015; Ucchesu et al., 2017) and in agronomy to distinguish and group cultivars (Orrù et al., 2012; Orrù et al., 2015; Piras et al., 2016; Sarigu et al., 2017; Sau et al., 2019; Sau et al., 2018). Images can be acquired using a digital camera or a flatbed scanner. Compared to a digital camera, a flatbed scanner has the advantages of coherent illumination and a known image scale, commonly expressed in dots per inch (DPI), as well as offering quality and speed of workflow execution (Lind, 2012).

The image acquisition stage involves creating the input images to be processed and making them available to the system, and represents the first stage of the analysis process used to obtain accurate measurements of the morpho-colorimetric features of the samples. The second stage in this procedure consists of pre-processing, which involves enhancing or recovering the image in order to emphasize its characteristics or regions of interest and thus obtain an optimised segmentation. The third stage, segmentation, consists of identifying the regions of interest in the image (foreground) and separating them from the background. This stage is one of the most complicated parts of the image processing method, as it is difficult to identify and separate uniform and homogeneous objects from the background (Scherer, 2020). In essence, the segmentation stage allows us to split the image into two parts, by identifying the edges of objects and separating them from the background (Chaki and Dey, 2018). Separating these objects of interest from the background is an important part of the analysis process, which allows us to count and measure the morpho-colorimetric features of each individual object by excluding the background, which is not of interest in terms of measurement.

The extensive literature cited in this paper relates to many botanical and archaeobotanical works in which image analysis has been used. In all these works, the method of acquisition (also called the double image-based method, DIM) of these images took place in two phases, via the acquisition of two images of the same seed samples: one with a white background, and the other with a black background. This operation allows us to obtain two overlapping images of the same sample during the pre-processing phase, with the aim of obtaining a single image to be segmented. This procedure is used in order to minimise image binarisation errors (Grillo, 2009), although it has certain disadvantages. The first is the possibility that during the acquisition phase of the two images, the seeds placed on the scanner may be moved, thus generating an overlapping error. In addition, the time taken to perform two image acquisitions of the same seed sample is longer than for a single image. A third drawback is the impossibility of processing a batch of images rather than a single image at a time.

For this reason, we propose a new single image-based method (SIM). The aims of this study are as follows:

- To develop a new open source plugin for the automatic segmentation of a single image acquisition;
- To test the segmentation performed by the new plugin on a single image, acquired with different blue-toned backgrounds;
- To compare the results obtained from the different segmentations, in order to identify the best procedure in terms of quality and execution time;
- To enable the processing of a batch of images in a unique workflow.

## 2. Materials and Methods

The method proposed in this paper was developed using ImageJ, an open source software for digital image processing designed by Wayne Rasband, National Institute of Health of the United States, in 1997. ImageJ is written in Java, and the complete source code is available to anyone who wants to make a contribution to the ImageJ community or simply to study it (Schneider et al., 2012). ImageJ allows the user to view, analyse, modify, process, save and print greyscale and colour images, and supports the most common image formats. ImageJ was designed with an open architecture that can be extended with plugins and recordable macros, and has a built-in plugin editor. The previous reasons allowed our research group to develop a plugin (Campanile, 2019). It allows for feature extraction from biological organisms, for use in carpology, a discipline that studies the seeds and fruits of spermatophytes from both a morphological and a structural point of view. When applied to remains from the past (paleocarpology), results of fundamental importance for paleobotany, paleoenvironmental and ecology studies can be generated. It was based on DIM, as defined in the previous section.

The use of image analysis techniques on seeds, rather than manual analysis, has several advantages; for example, it speeds up the analysis process, minimises distortions created by natural light and microscopes, and enables some characteristics to be automatically identified based on image pixel values.

In order to test our automatic segmentation method, an image database was built using a total of 3,386 samples of seeds from 120 plant species belonging to the Fabaceae family (SM1). This family was chosen because it is one of the largest and has great variability in terms of both the size and colour of the seeds. All samples were drawn from the base collection in the Sardinian Germplasm Bank (BG-SAR), University of Cagliari, Italy.

## 2.1. Image Database

For each different sample of seeds from the plant species shown in Table 1, seven images were captured for the image database, giving a total of 720 digital images containing 20,316 seeds. These images were subdivided into two groups, G1 and G2. For the first group (G1), 120 seed samples were scanned, each with four different hues of blue in the background, giving a total of 480 digital images containing 13,544 seeds. The hues used were drawn from the RGB colour model (R – red, G – green and B – blue), and were used to define the subgroups of G1, as follows: G1.1 contained the light blue images (R E 140, G E 220, B E 250), G1.2 contained the medium-light blue images (R E 25, G E 130, B E 210), G1.3 contained the medium-dark blue images (R E 20, G E 60, B E 150) and G1.4 contained the dark blue images (R E 30, G E 60, B E 100). For the same 120 samples of seeds, a second group (G2) was created with two different tones in the background, black (R E 0, G E 0, B E 0) and white (R E 255, G E 255, B E 255), giving a total of 240 images containing 6,772 seeds.

The digital images in both groups G1 and G2 were acquired using an Epson Perfection V550 flatbed scanner.

An example of a single sample of *Albizia lophanta* containing 13 seeds, prepared for groups G1 and G2, can be seen in Figure 1.

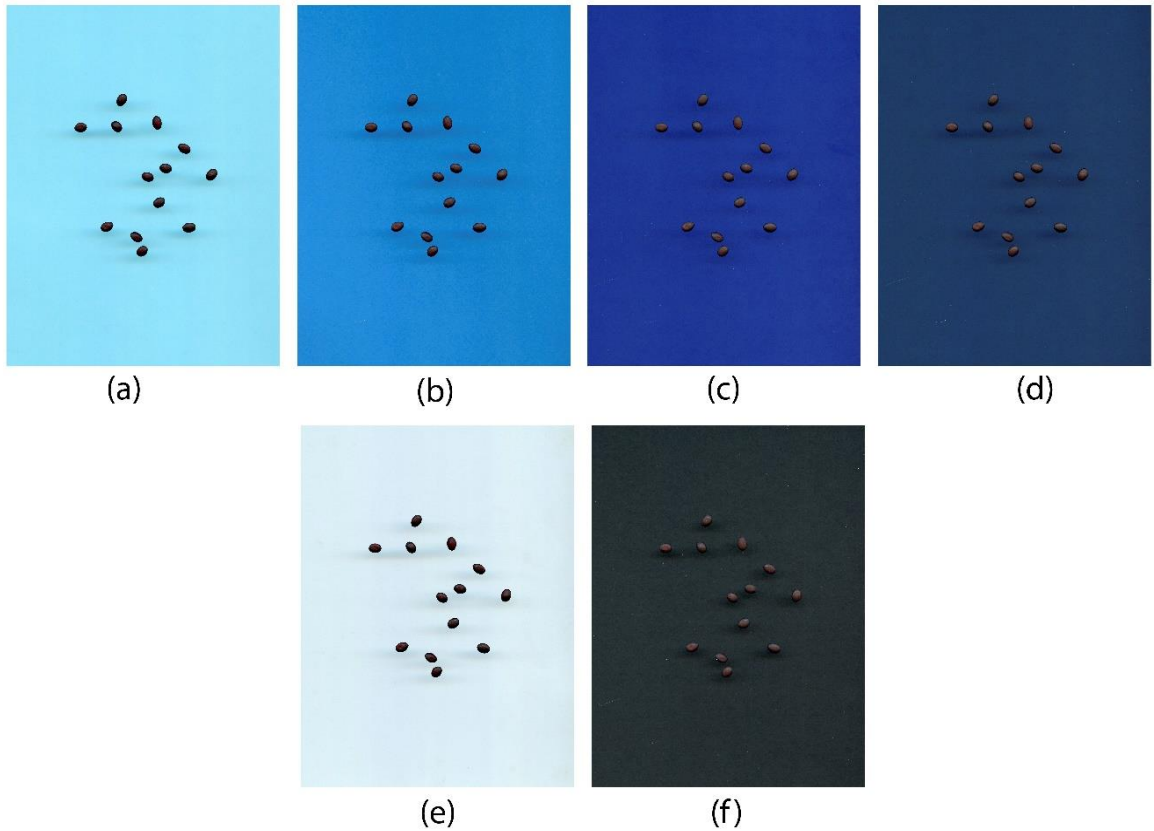

**Figure 1.** Digital images of *Albizia lophanta* processed for G1: (a) G1.1 light blue background; (b) G1.2 medium-light blue background; (c) G1.3 medium-dark blue background; (d) G1.4 dark blue background; and G2: (e) white background; (f) black background.

## 2.2. Proposed Process: Acquisition Stage

During acquisition, seeds should be arranged on the flatbed scanner separately from each other, in order to avoid overlap (Figure 2).

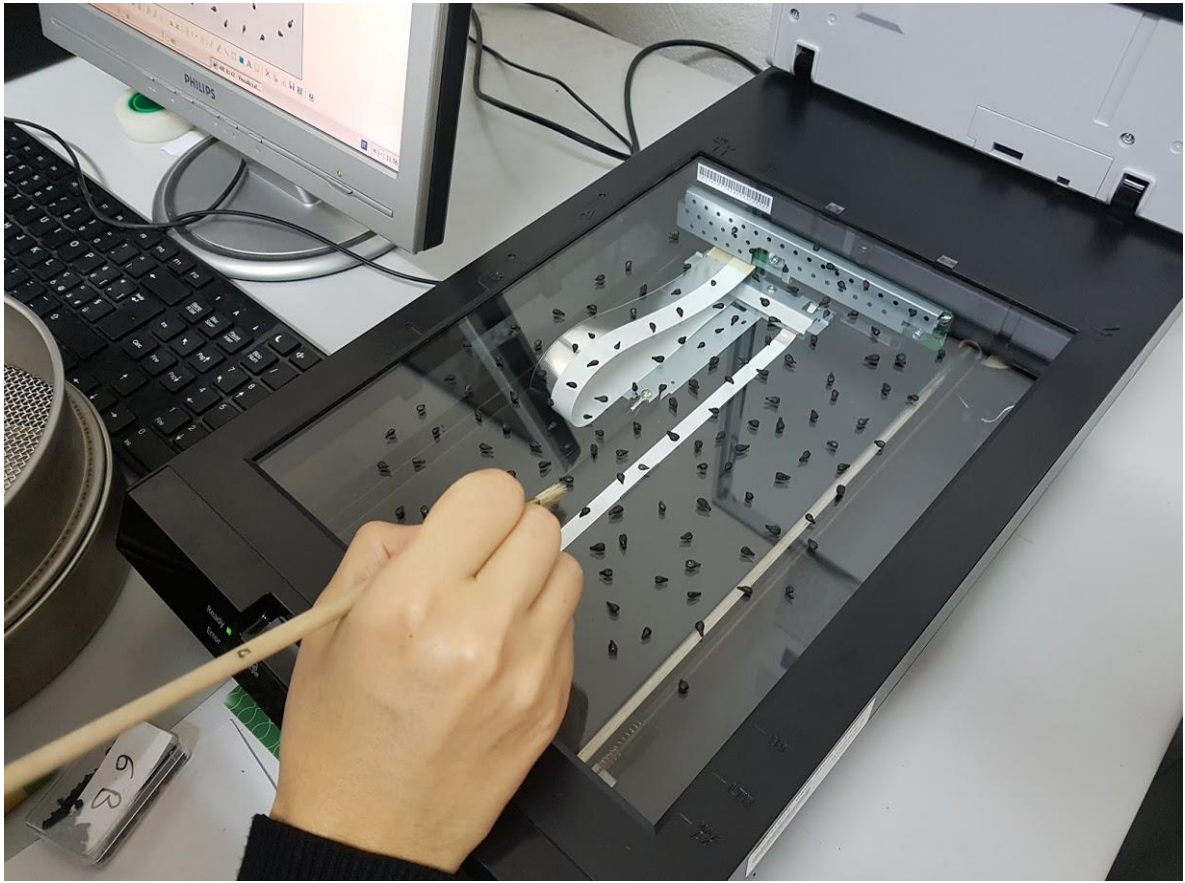

**Figure 2.** Arrangement of seeds on the flatbed scanner.

The area occupied by the seeds is then covered with a tray lined with a blue background for the digital image. The shade of blue is not fixed, and varies from light to dark blue; the variation in tone does not alter the segmentation results. Finally, the acquisition process is performed using a minimum resolution of 400 dpi, and the resulting image is saved in JPEG format.

Since the subsequent stages involve batch processing, in which a group of images are processed at a time, all the images to be analysed must be stored in a common folder.

## 2.3. Proposed Process: Pre-Processing Stage

In this stage, the region of interest of the problem is initially defined by the set of seeds to be analysed in each image. In order to optimise the segmentation stage, RGB images are converted to grayscale, and edge detection is performed for each original image previously acquired (Figure 3).

After selecting the folder in which the batch of images is located, our algorithm is applied to each image as follows:

- Step 1: Open the original RGB image;
- Step 2: Convert the RGB image to a greyscale image;
- Step 3: Apply edge detection to the image.

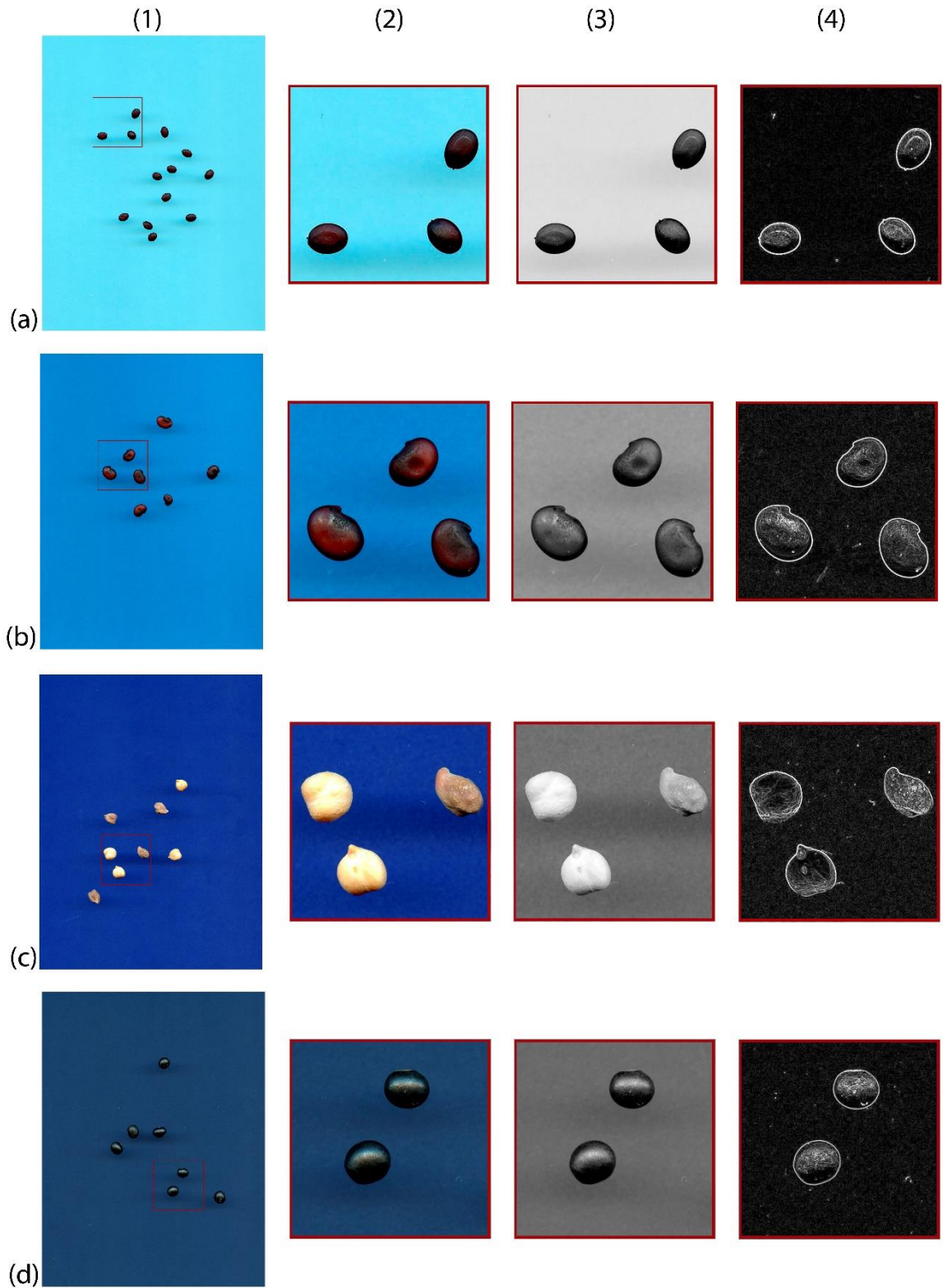

**Figure 3.** Example of results from each step of the pre-processing stage for some of the original images: (1) original image; (2) crop of the original image; (3) conversion to greyscale; and (4) edge detection. Images are (a) G1.1 *Albizia lophanta*; (b) G1.2 *Adenocarpus complicates*; (c) G1.3 *Lathyrus ochrus*; and (d) G1.4 *Glycine max*.

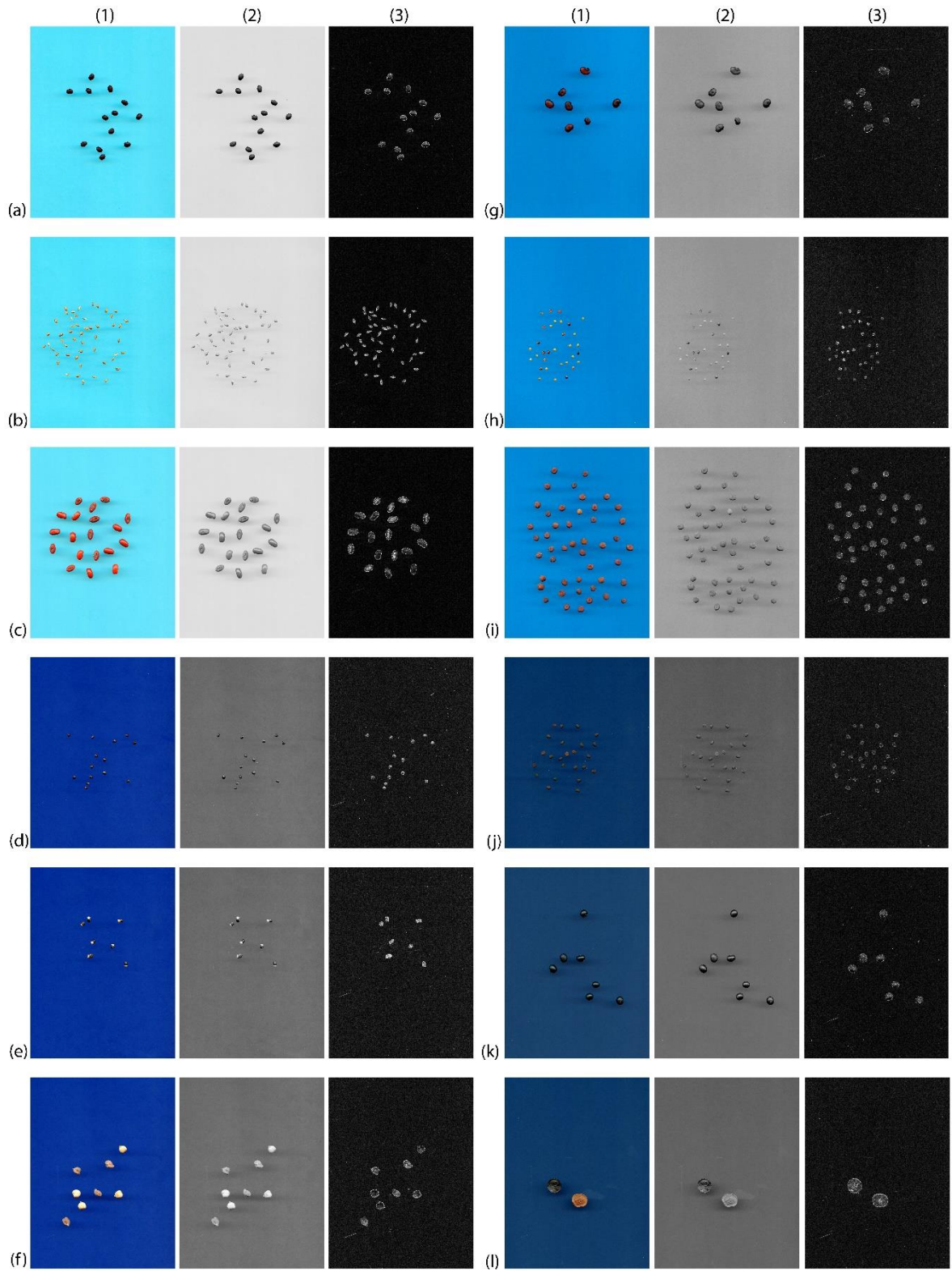

**Figure 4.** Final results from each step of the pre-processing stage for several samples, scanned with different shades of blue in the background: (1) pre-processing of the original image; (2) greyscale image; and (3) edge detection. Images are of (a) G1.1 *Albizia lophanta*;

(b) G1.1 *Anthyllis barba-jovis*; (c) G1.1 *Erythrina caffra*; (d) G1.2 *Astragalus onobrychis*; (e) G1.2 *Calicotome villosa*; (f) G1.2 *Lathyrus ochrus*; (g) G1.3 *Adenocarpus complicatus*; (h) G1.3 *Chamaecytisus proliferus*; (i) G1.3 *Cicer arietinum*; (j) G1.4 *Colutea media*; (k) G1.4 *Glycine max*; and (l) G1.4 *Bauhinia variegata*.

## 2.4. Proposed Process: Segmentation stage

After the pre-processing step, the segmentation stage is applied, which consists of a set of operations that subdivide each original image into two distinct regions: the foreground (regions of interest, i.e. seeds) and the background. More precisely, for each pre-processed image and its respective original RGB image, the segmentation stage involves identifying and removing the blue background using the RGB and HSB (H – hue, S – saturation, B – brightness) colour models, region filling and noise correction.

The algorithm for this stage can be briefly described as follows:

- Step 1: Identify and remove pixels in shades of blue. After converting the input image from the RGB colour space into the HSB colour space, create a new binary image by removing only the pixels that satisfy the following conditions:
    - The intensity of channel B is the highest, and there is a difference between the intensities of channels B and R;
    - The intensity of the H channel reflects a blue tone.
- Step 2: Apply the region filling operation;
- Step 3: Remove objects (considered to be noise) with an area less than 1/3 of the largest area found.

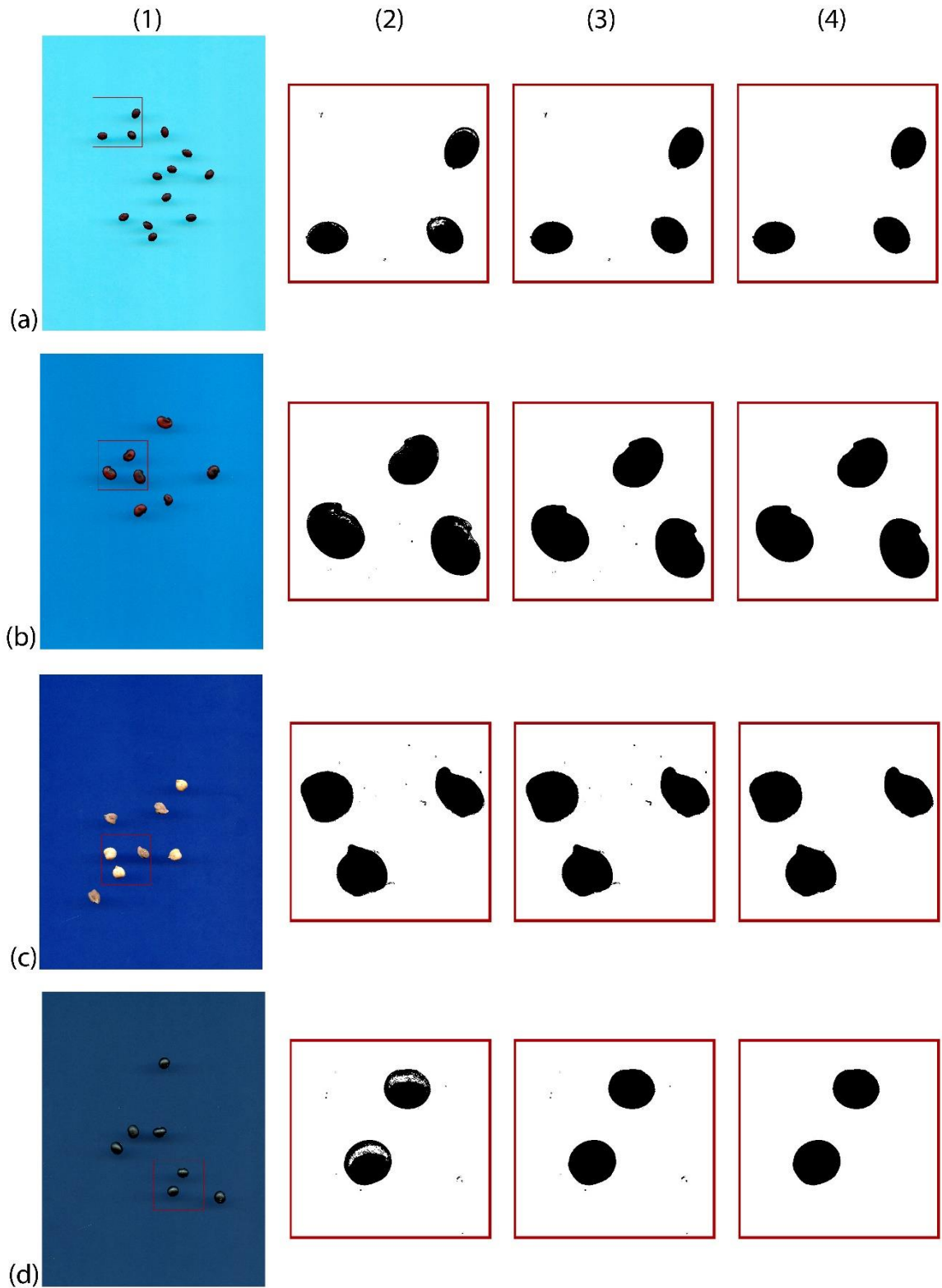

**Figure 5.** Example results from each step of the segmentation stage: (1) original images; (2) binary images of the cropped area, resulting from the removal of shades of blue background pixels; (3) region filling; and (4) noise removal. Images are of (a) G1.1 *Albizia lophanta*; (b) G1.2 *Adenocarpus complicatus*; (c) G1.3 *Lathyrus ochrus*; and (d) G1.4 *Glycine max*.

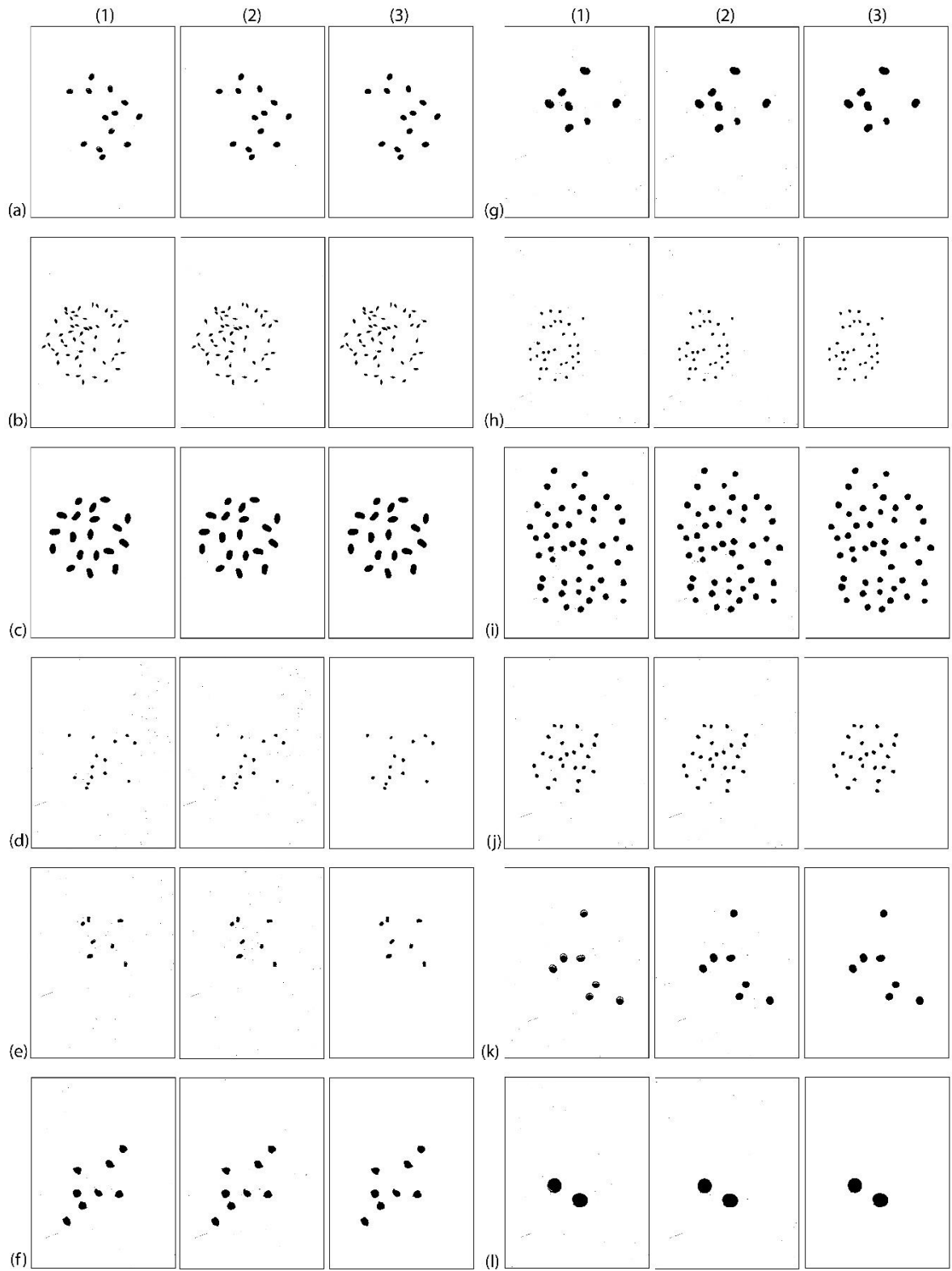

**Figure 6.** Final results of each step of the segmentation stage, for several samples scanned with different shades of blue in the background: (1) binary image resulting from the removal of pixels in shades of blue; (2) region filling; and (3) noise removal. Images are of (a) G1.1

*Albizia lophanta*; (b) G1.1 *Anthyllis barba-jovis*; (c) G1.1 *Erythrina caffra*; (d) G1.2 *Astragalus onobrychis*; (e) G1.2 *Calicotome villosa*; (f) G1.2 *Lathyrus ochrus*; (g) G1.3 *Adenocarpus complicatus*; (h) G1.3 *Chamaecytisus proliferus*; (i) G1.3 *Cicer arietinum*; (j) G1.4 *Colutea media*; (k) G1.4 *Glycine max*; and (l) G1.4 *Bauhinia variegata*.

## 2.5. Acquisition times: DIM vs. SIM

Since the workflow starts with the acquisition of the image to be analysed, it is clear that the time spent in this stage is important, since it affects the total time consumed by an automatic image analysis system. The acquisition time refers to the period between the placement of the seeds on the digitiser table of a previously configured scanner and the recording of the digital image file. The purpose of this test is to compare the average acquisition time for the two input image patterns recorded with the DIM approach (i.e. using images with black and white backgrounds, with two images to be processed for each sample) and the SIM approach (i.e. using images with a blue background, with one image for each sample to be processed). In both cases, a timer was started when the operator began to place the sample seeds on the digitising table, and was stopped when the final file had been recorded. For this test, we used the same scanner, configured with a standard resolution of 400 DPI, and images were recorded with a scanning area of 2,125 x 2,834 pixels in JPEG 2.5.2 format.

## 2.6. Runtime test: DIM vs. SIM

In order to compare the time spent on the execution of the two different methods, from the opening of the previously digitised images to the final processing results, the execution time was measured. These measurements were made considering their individual outlines. Since DIM requires two different images (with white and black backgrounds) for a single segmentation, the total execution time was considered to be the sum of the execution times for all images in the G2 group from the image database. Since SIM involves the opening and processing of a batch for each execution, the total processing time was considered to be the time spent on processing the entire batch of images in the image database, for each subgroup of the G1 group.

2.7. Seed count and segmentation test: DIM vs. SIM

In order to evaluate the seed counts for each sample in the image bank, and to evaluate possible segmentation errors, the automatic seed counts of the images segmented using DIM and SIM were compared with the manual seed counts for the original images. For each method, the percentages of equal counts (successes) and different counts (errors) were calculated and compared to the manual counts.

3. Results

The new features offered by SIM, such as the possibility of acquiring a single image per sample, the use of backgrounds in various shades of blue, and the use of batch image processing, improved the quality of the segmentation and gave a significant reduction in processing time.

3.1. Acquisition and runtime tests: DIM vs. SIM

The time taken to place an average of 100 seeds on the scanner and to acquire them with a blue background (SIM) was recorded as 59.9 s. In contrast, the time required for positioning and image acquisition using a black and white background (DIM) was 96.5 s. The time required for segmentation of a single image (SIM) was 0.02 s, as compared with 63 s for the double image (DIM).

3.2. Seed count and segmentation test: DIM, SIM and manual method

No errors in image segmentation were detected for SIM (Figs. 5 and 6). The new plugin was able to correctly separate the blue background from the seeds, removing all particles that were not seeds. In contrast, with DIM, 26 samples showed errors in segmentation due to overlapping of the seeds, generated during image acquisition (Figure 7).

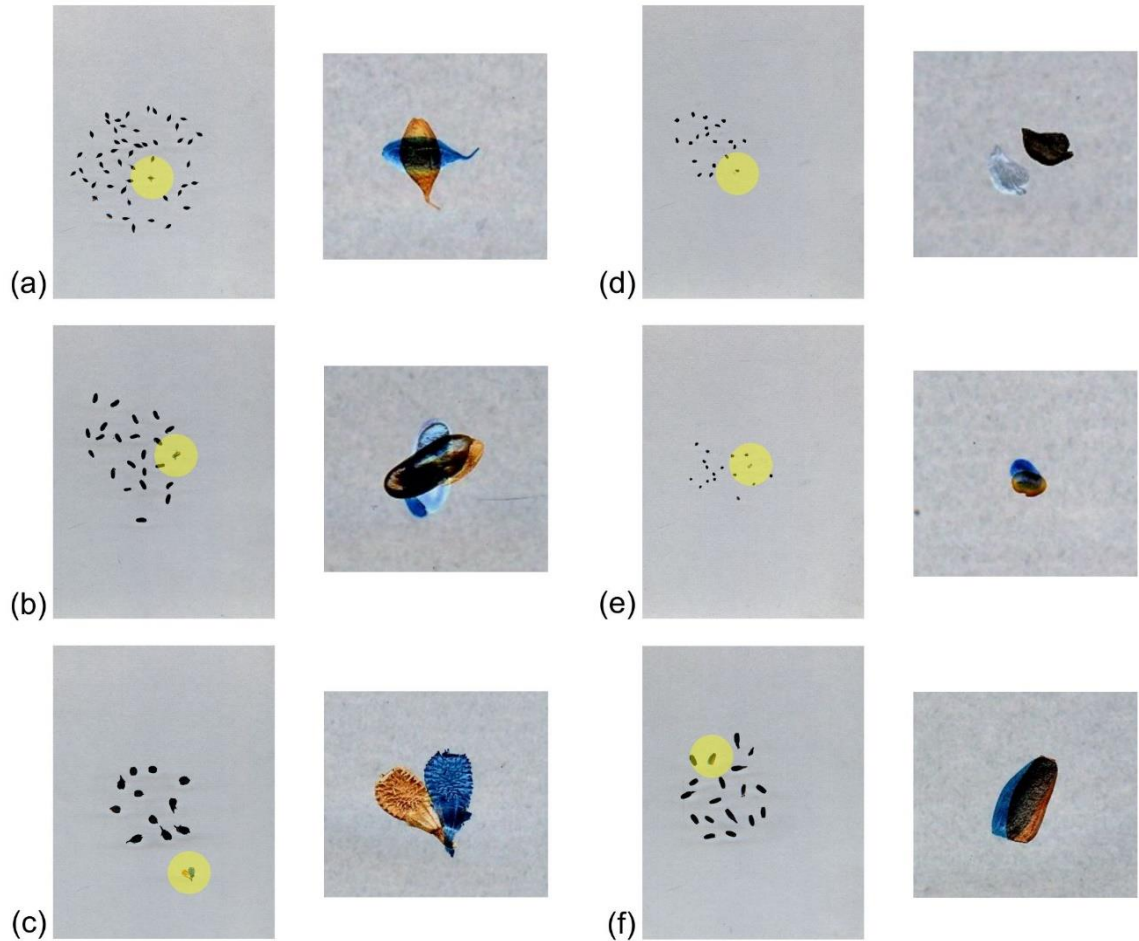

**Figure 7.** Four of the twenty-six cases of incorrect image overlapping using DIM: (a) *Anthyllis barba-jovis*; (b) *Acacia saligna*; (c) *Hedysarum coronarium*; (d) *Erophaca baetica*; (e) *Astragalus penduliflorus*; and (f) *Coronilla valentina*

## 4. Discussion

Precise and accurate measurements of the morphocolorimetric features of the seeds through image analysis are essential in order to obtain reliable data for use in taxonomic studies and plant classification. In this work, we compared two methods of acquiring seed images and developed a new plugin for image segmentation. Four different groups of images were used in this study in order to demonstrate the validity of the proposed SIM plugin, from acquisition to the pre-processing and segmentation of seed samples. Based on the reduction in the acquisition and segmentation times for the images, and considering the same count of elements after this process, SIM is proven to be more efficient than DIM in terms of both the response time and the requirements and sample preparation for the acquisition of images.

During the image acquisition process, both SIM and DIM require the same care from the operator in terms of the cleaning of the sample and separation of the sampled seeds. However, while DIM requires two images to be acquired for each sample, SIM only requires the digitisation of one image. Thus, SIM provides a considerable reduction in the time allocated to this process. In addition, the tests performed on SIM using four different shades of blue, ranging from the lightest to the darkest, shows that this approach is able to work perfectly with any shade. The operator can therefore freely choose any shade of blue for the background, without constraints.

In terms of the sensitivity of the image acquisition process, DIM requires that for a given sample, two images should be acquired with no change in the positions of the seeds. This is because if the positions of the seeds undergo any changes, overlapping errors are generated in the segmented image, causing changes in the shape of the seeds and consequently errors in the final results. This sensitivity means that the system operator must be very careful to keep the seed positions unchanged during the acquisition process, in order to avoid the need to abandon the images and repeat the entire process. In the SIM method, this is not necessary.

The SIM method is also more efficient in terms of runtime, as it allows for the processing of a batch of images, rather than individual images as in DIM. Thus, the operator is only responsible for choosing the directory containing the batch of images, rather than a single image at a time (DIM).

After acquisition, pre-processing and segmentation of the samples, both SIM and DIM gave results that were identical to those observed in manual segmentation. Hence, regarding the seed counts of the segmented images, both automatic methods were equally efficient.

## 5. Conclusions

The open source plugin proposed here is proven to be highly efficient in terms of both time and segmentation when working with large numbers of images and a wide diversity of seed shapes.

SIM can therefore be used to achieve reliable seed segmentation, within a short time and with minimal effort by the operator. Consequently, we can easily extract any desired morpho-colorimetric information. The plugin is available to download at[1].

[1]https://github.com/andrealoddo/blueBackgroundSeedsSegmenter

## Acknowledgements

The Regione Autonoma della Sardegna partially supported the research in this paper as part of a research project entitled "Algorithms and Models for Imaging Science [AMIS]" (finanziato con risorse FSC 2014--2020 Patto per lo Sviluppo della Regione Sardegna).